\pdfoutput=1

\documentclass[11pt]{article}
\usepackage{acl_arr}

\usepackage{times}
\usepackage{latexsym}

\usepackage[T1]{fontenc}

\usepackage[utf8]{inputenc}
\usepackage{microtype}

\usepackage[english]{babel}
\usepackage[autostyle]{csquotes}
\usepackage{algorithm}
\usepackage[noend]{algpseudocode}
\usepackage{varwidth}
\usepackage{multirow}
\usepackage{tabularx}
\usepackage{subcaption}
\usepackage{amssymb}
\usepackage{amsmath}
\usepackage{amsthm}
\usepackage{amsfonts}
\usepackage{bbm}
\usepackage{bm}
\usepackage{booktabs}
\usepackage{graphicx}
\usepackage{multirow}
\usepackage{multicol}
\usepackage{enumitem}
\usepackage{mdwlist}
\usepackage{csquotes}
\usepackage{xcolor}
\setlist{nosep}

\title{InteractiveIE: Towards Assessing the Strength of Human-AI Collaboration in Improving the Performance of Information Extraction}

\author{Ishani Mondal$^{1}$, Michelle Yuan$^{6}$,   
 \textbf{Anandhavelu N}$^{2}$, \textbf{Aparna Garimella}$^{2}$, \\
 \textbf{Francis Ferraro}$^{3}$, \textbf{Andrew Blair-Stanek}$^{5}$, \textbf{Benjamin Van Durme}$^{6}$, \textbf{Jordan Boyd-Graber}$^{1}$ \\
  $^{1}$ University of Maryland, College Park, \hspace{0.2cm} 
  $^{2}$ Adobe Research, India \hspace{0.2cm}, \\
  $^{3}$ University of Maryland, Baltimore County, \hspace{0.2cm} 
  $^{4}$ University of Maryland School of Law, \hspace{0.2cm} \\ 
  $^{5}$ John Hopkins University, \hspace{0.2cm}
  $^{6}$ Amazon, New York\\
  \texttt{imondal@umd.edu}\\
}

\date{}

\newif\ifcomment\commenttrue

\usepackage[a-1b]{pdfx}

\usepackage{framed}
\usepackage{mdwlist}
\usepackage{siunitx}
\usepackage{latexsym}
\usepackage{colortbl}
\usepackage{xcolor}
\usepackage{nicefrac}
\usepackage{booktabs}
\usepackage{fnpct}
\usepackage{amsfonts}
\usepackage[T1]{fontenc}
\usepackage{bold-extra}
\usepackage{amsmath}
\usepackage{amssymb}
\usepackage{bm}
\usepackage{graphicx}
\usepackage{mathtools}
\usepackage{microtype}
\usepackage{multirow}
\usepackage{multicol}
\usepackage{xpatch}
\usepackage{latexsym,comment}
\usepackage[normalem]{ulem}

\newcommand*{\missingreference}{{\Huge \colorbox{red}{?reference?}}}
\newcommand*{\missingcitation}{{\Huge \colorbox{red}{?citation?}}}

\makeatletter
\xpatchcmd{\@setref}{\bfseries}{\missingreference}{}{}
\def\@citex[#1]#2{\leavevmode
    \let\@citea\@empty
    \@cite{\@for\@citeb:=#2\do
        {\@citea\def\@citea{,\penalty\@m\ }%
            \edef\@citeb{\expandafter\@firstofone\@citeb\@empty}%
            \if@filesw\immediate\write\@auxout{\string\citation{\@citeb}}\fi
            \@ifundefined{b@\@citeb}{\hbox{\reset@font\missingcitation}%
                \G@refundefinedtrue
                \@latex@warning
                {Citation `\@citeb' on page \thepage \space undefined}}%
            {\@cite@ofmt{\csname b@\@citeb\endcsname}}}}{#1}}
\makeatother

\newcommand{\gem}[1]{\mbox{\textsc{gem}}}

\newcommand{\hidetext}[1]{}
\newcommand{\ignore}[1]{}

\ifcomment
    \newcommand{\pinaforecomment}[3]{\colorbox{#1}{\parbox{.8\linewidth}{#2: #3}}}

    \newcommand{\prtodo}[1]{\pinaforecomment{lightblue}{pr}{#1}}
    \newcommand{\prtodoi}[1]{\pinaforecomment{lightblue}{pr}{#1}}
\else
    \newcommand{\pinaforecomment}[3]{}
    \newcommand{\prtodo}[1]{}
    \newcommand{\prtodoi}[1]{}
\fi

\newcommand{\smallurl}[1]{ \begin{tiny}\url{#1}\end{tiny}}

\definecolor{lightblue}{HTML}{3cc7ea}
\definecolor{CUgold}{HTML}{CFB87C}
\definecolor{grey}{rgb}{0.95,0.95,0.95}
\definecolor{ceil}{rgb}{0.57, 0.63, 0.81}
\definecolor{UMDred}{HTML}{ed1c24}
\definecolor{UMDyellow}{HTML}{ffc20e}

\usepackage[normalem]{ulem}

\graphicspath{ {figures/}{auto_commit_fig/}{auto_fig/} }

\begin{document}

\maketitle
\begin{abstract}
Learning template-based information extraction (IE) from documents is a crucial yet difficult task.
Prior template-based IE approaches assume foreknowledge of the domain’s templates. 
However, many real-world IE scenarios do not have pre-defined schemas. To "figure-out-as you go" requires a solution with zero or minimal prior supervision. 
To quickly bootstrap templates in a real-world setting, we need to induce template slots from the documents with zero or minimal supervision.
To address the above needs, we introduce \textbf{InteractiveIE},
a human-in-the-loop interactive interface where initially questions are automatically generated from entities in the corpus, followed by explanation-driven clustering of these questions, then allowing the users to modify, add, or otherwise edit questions based on their specific information needs. 
Besides, we provide agency to the humans at intermediate steps such as: tweaking the automatically generated questions followed by re-arranging those in different clusters to generate schema.
After conducting empirical human study, we observe that there is a gradual improvement of information mapping to desired slots using \textit{InteractiveIE} over \textit{AI-only} baseline 
with minimum number of interactions with the interface. Our method has been shown to be easily extensible to new domains (biomedical or legal), where procuring training data is expensive. 
Furthermore, we observe that explanations provided by clustering model fairly helped to guide the users in making sense of IE schema over time.
\end{abstract}

  
  \section{Introduction}
  \label{sec:intro}
  The goal of information extraction (IE) is to learn structure from unstructured documents. 
Existing IE tools help the analysts understand certain patterns or behaviors in the world \cite{li-etal-2022-document, mora-etal-2009-exploring}. 
However, in a fast-moving real-world situation, IE needs are likely to change over time, and it is not possible to have prior knowledge of required information intent beforehand. 
For instance, during a pandemic, people might be concerned about the mortality rate of a disease at some point, whereas they might be interested in knowing the immunization steps at a later time
Perhaps, people might need to extract information like \textbf{mortality rate in Mexico} during the onset of pandemic before planning for a travel in 2020 beginning, and after 2021, they would be more interested to extract \textbf{vaccines near Mexico city}
Despite this ever-growing need, widely popular supervised IE systems rely on human annotations of templates \cite{chinchor-marsh-1998-appendix,pavlick-etal-2016-gun}. 

In such a case, an unsupervised approach would be ideal for quickly building IE systems in new domains where models can learn the type of information without relying on expensive annotation of documents for training purposes. 
Prior unsupervised approaches are mainly probabilistic from modeling patterns in clauses \cite{bamman-smith-2014-unsupervised}. 
Still, template matching accuracy is quite low for these methods.

\begin{figure}[!t]
\fbox{\includegraphics[width=0.45\textwidth]{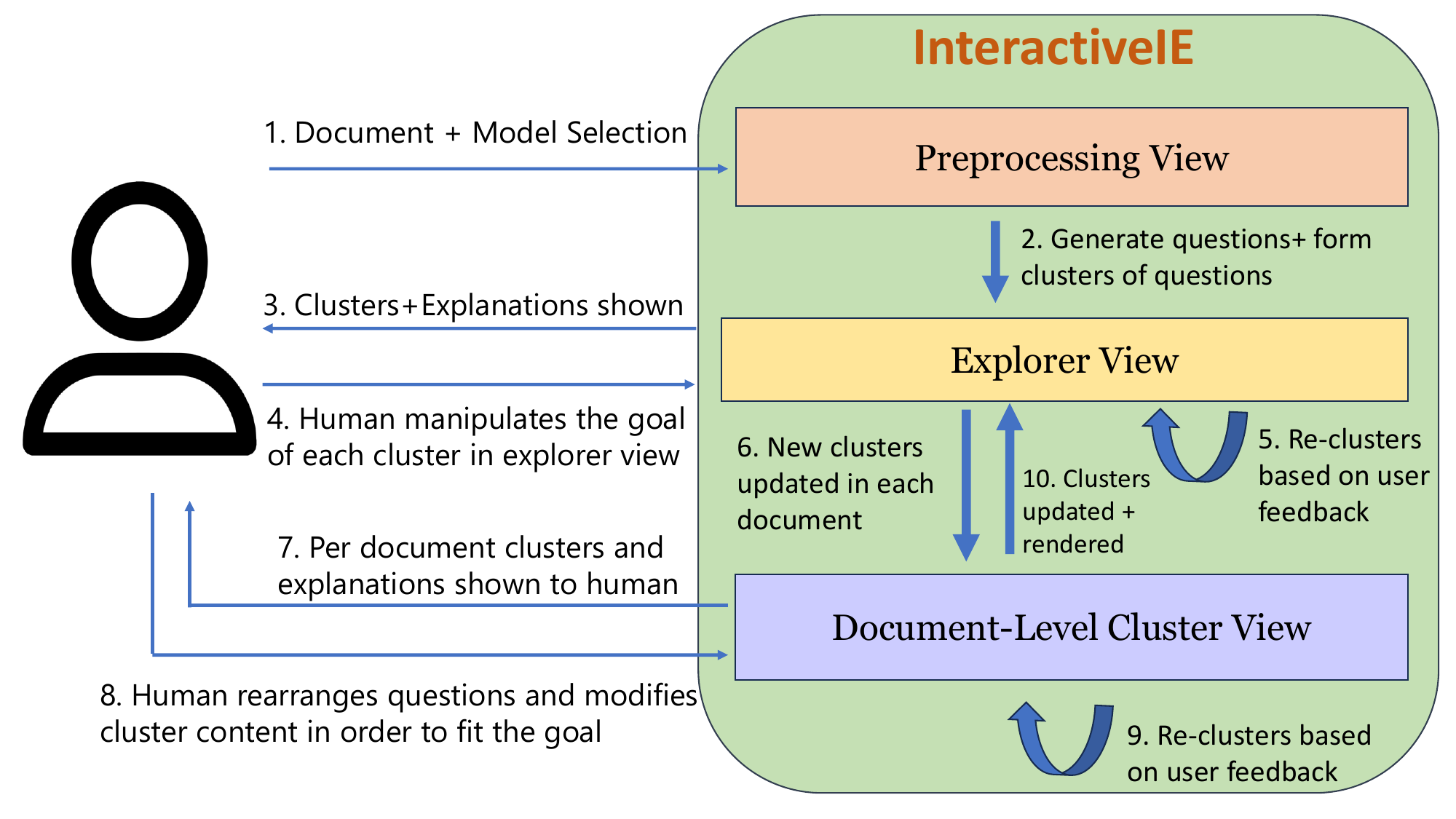}}
\caption{shows Human-AI interactions in \textit{InteractiveIE} which consists of three main components: Preprocessing View, Explorer View and Document-Level Cluster view. Through this interface, the humans would be able to modify the bird's eye view of a corpus by altering the questions and reclustering those.}
\label{fig:humanaidiagram}
\end{figure}

\begin{figure*}
\includegraphics[width=\linewidth]{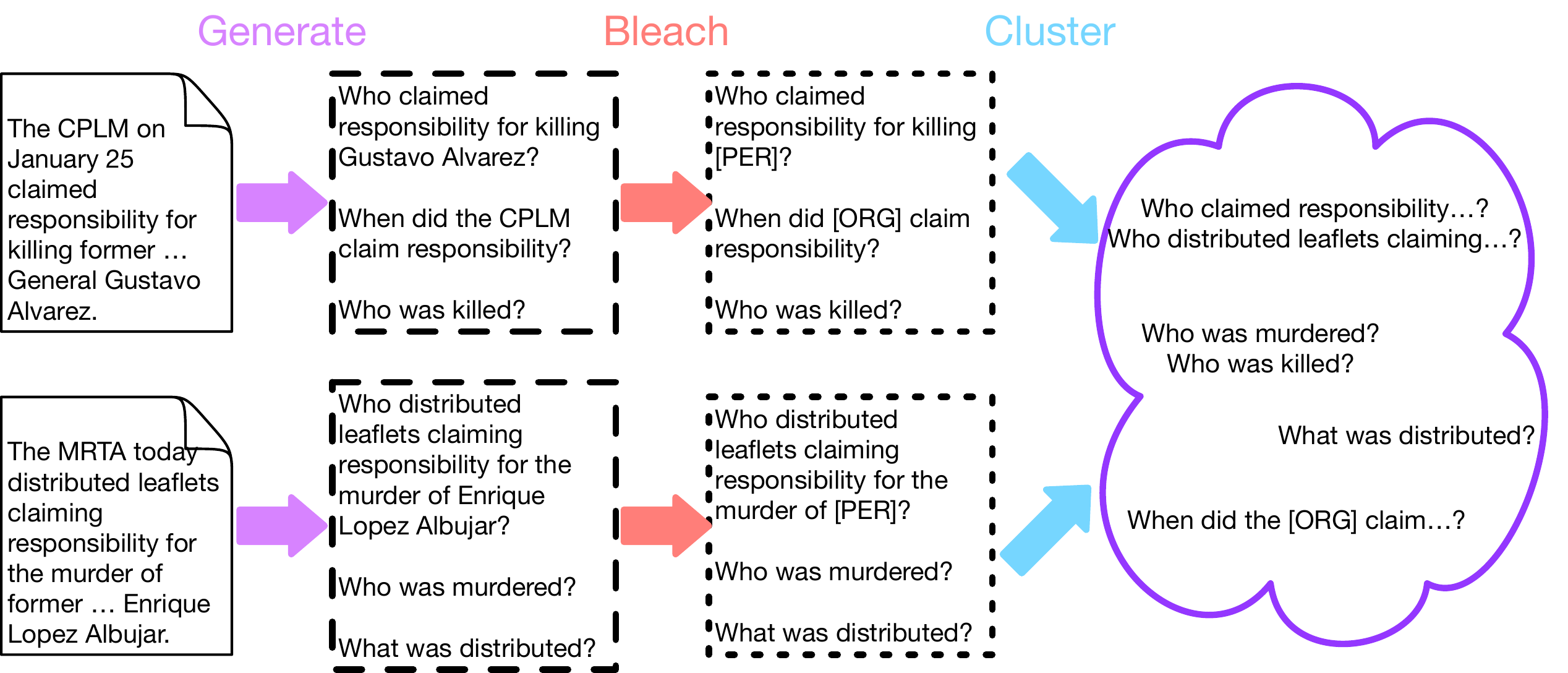}
\caption{Inducing the important relationships and events from a corpus can be performed using question generation from the corpus since questions are representatives of information need. This figure shows the process of generating factoid questions from a text corpus, then modifying these questions by substituting entities with Named Entity Recognition (NER) tags. This process results in bleached questions, which we then embed and cluster into categories corresponding to different schema slot types. The accompanying illustration demonstrates this approach, showcasing the transition from original questions to bleached versions and their subsequent organization into relevant groups.}
\label{fig:schemamethod}
\end{figure*}

A quick way of defining an information need is by asking a question.
Recent work in question generation can automatically output factoid questions conditioned on a given passage \cite{nagumothu-etal-2022-pie, promptore}. Factoid questions typically ask which entity in the passage fulfills a specified semantic role. For instance, if the input is a legal contract, the model can generate questions about the time of agreement, how long would that be effective and who all have agreed upon the contract. In the same example, we can map those generated questions into specific semantic labels such as, \enquote{Agreement Date}, \enquote{Effective Date}, \enquote{Agreed upon Parties} respectively. 

Therefore, in this paper, we first propose an unsupervised method of looking at question generation using state-of-the-art models like T5 \cite{Raffel2019ExploringTL}, BART \cite{lewis-etal-2020-bart}  to induce slots. We cluster the generated questions to find a collection of representative questions that should adhere to slot types that are present across documents. 
We have used real-human experiments in this work to assess whether explanation-driven clustering algorithm can indeed guide the users to refine their own schema over time and satisfy their dynamic information needs. 
Through \textbf{InteractiveIE}, we allow the users to perform the following operations in three different modes: \textit{a) Preprocessing View}: Users run \textit{UnsupervisedIE} on some chosen documents (Steps 1, 2 in \ref{fig:humanaidiagram}), \textit{b) Clustering Explorer View}: Users view the clusters and explanations of generated clusters and can modify the explanations to extract desired information from documents (Steps 3, 4, 5, 6 in \ref{fig:humanaidiagram}) and \textit{c) Document-Level Cluster view}: users are allowed to re-arrange questions among the clusters, modify questions by reading the corresponding document (Steps 7, 8, 9, 10 in \ref{fig:humanaidiagram}). 
We evaluate the effectiveness of our interface through human studies and observed that with a minimal human supervision, slot mapping performance improves over unsupervised baselines to a considerable amount.

\section{Background on Schema Induction}
\label{sec:autoschema}
Unsupervised RE (or OpenRE \cite{yates-etal-2007-textrunner}) aims to extract relations without having access to a labeled dataset during training. Yates et al. \cite{yates-etal-2007-textrunner} extract triple candidates using syntactic rules and refine the candidates with a trained scorer. Saha et al. \cite{saha-mausam-2018-open} propose to simplify conjunctive sentences to improve triples extraction. More recently, neural networks and word-embedding were applied to solve this task \cite{cui-etal-2018-neural}, requiring a general domain annotated dataset to pretrain their model. Finally, Roy et al. \cite{roy-etal-2019-supervising} propose an ensemble method to aggregate results of multiple OpenRE models. These triples extraction approaches rely on surface forms, which makes it hard for them to group instances that express the same relation using very different words and syntax. 

Although the existing OpenIE based models \cite{hu-etal-2020-selfore,renze-etal-2021-unified,marcheggiani-titov-2016-discrete,tran-etal-2020-revisiting} extract relations from unannotated datasets, we argue that they are not truly unsupervised approaches. The main problem is hyperparameter tuning (these methods rely extensively on hyperparameters that need to be adjusted i.e. the number of epochs, regularization and learning rate). All of these can only be determined from chunks of training data. However, in a real-world scenario it is very difficult to estimate them without access to enough labeled data. Therefore, we argue that the previous methods are not fully unsupervised when it comes to hyperparameter tuning, which we believe, restricts their application in a real-world setting. Recently, \cite{promptore} has proposed an unsupervised method by encoding the relations between entities using a relation encoder followed by clustering the relation representations. However, they fix the entities before performing such operation, which is also a bit unrealistic. \emph{As a result, it motivates us to define the unsupervised IE setting as learning an IE model and tuning its hyperparameters using unannotated data on-the-fly}.

Some recent works on event schema generation used transformers to handle schema generation in a complex
scenario \cite{li-etal-2020-connecting,li-etal-2021-future}. However, this approach was
unable to transfer to new domains where the supervised event retrieval and extraction model failed. Thus, neither do they offer a perfect solution for schema induction without manual postprocessing, nor build a timely human correction
system \cite{du-etal-2022-resin}. More recently, \cite{zhang-etal-2023-human} proposed an event schema generation using GPT-3 generated output, but this method might not be scalable for complex domains like biomedical or legal, since a complete GPT-3 based data generation might be hallucinating (hence, not reliable) in nature. In order to address this limitation, we propose a human-in-the-loop approach involves generated subject-verb-object tuples and question-answer pairs to induce events and relations between the entities from corpora of complex domains.

\section{UnsupervisedIE Methodology}
In case of supervised IE, the fixed templates and slots are annotated, whereas our goal is to extract information from documents dynamically on-the-fly without knowledge of pre-defined templates. For capturing the IE needs that change over time, we aim to define a way to quickly bootstrap template schemas with zero to minimal supervision. Since the goal of asking a question nicely intersects with defining an information need \cite{srihari-li-2000-question}, we use widely used Question-Answering systems to refine our dynamic schema. In this section, we define our method of determining templates for documents automatically (\textit{UnsupervisedIE}). Our pipeline of automatic template induction comprises of the following steps as described below: 

\subsection{Salient Entity Identification} To the potential entities (which might be possible slot-fillers), we extract both general domain and domain-specific named entities from the corpora.

\subsection{Generating Questions} We propose to use question generation to induce templates.
With advances in neural text generation~\citep{lewis-etal-2020-bart}, generating factoid questions
has become much easier ~\citep{lewis-etal-2021-paq}. The models generate questions based on a
context passage and an entity mention from the passage.
Each generated question inherently describes the information need of an entity mention in the document.
Using the same example from Figure~\ref{fig:schemamethod} 
, a model may generate the following question about ``MRTA'': ``who distributed leaflets claiming the responsibility for the murder of former defense minister Enrique Lopez Albujar''. If we represent ``MRTA'' with this generated question, we link it to other entities with generated questions that ask ``who claimed the responsibility for the murder''.
A cluster of these questions naturally maps to the \textit{Perpetrator} slot type. This motivates a question-driven approach to slot filling.

\subsection{Embedding Questions}
We embed the generated questions for representing them to cluster. 

\subsection{Explanation-Driven Clustering of Questions:}
After embedding the questions, we use two methods as described below to obtain the clusters.

\paragraph{A. Clustering with $k$-means}
\label{subsec:kmeans-cluster}
After clustering questions, we generate representative questions for each cluster (The goal is that each cluster can correspond to some slot type.  Then we determine the representative questions for each cluster. Based on cosine similarity among the questions, we select the ones which have the highest average similarity with other questions and choose top-k questions. For each document, we consider the document specific questions having high cosine similarity with the mean embedding of clusters as “representative” questions of that cluster in the document.), get explanations/rationales behind each cluster and an abbreviated description for each cluster. Based on the questions present in each cluster, we prompt gpt3.5 (ChatGPT) to generate free-form explanations (both long and short/crisp) behind why those questions have been clustered together using prompt \ref{appendix:promptA}.

\paragraph{B. Clustering using LLM}
\label{subsec:LLM-cluster}
Similar to $k$-means, here the questions from all the user-selected documents are clustered into $K$ groups using ChatGPT and asking for an explanation for why did it think of clustering the questions in that way (we have used the prompt \ref{appendix:promptB}).

\paragraph{Slot Mapping and Evaluation:}
\label{par:evaluation}
After running \textit{UnsupervisedIE} pipeline on documents, one can see the clusters of generated questions along with the gold slot if mapped already, otherwise they will only see the abbreviated description of each generated cluster (as described in the previous section). Here we describe, how does the output of the \textit{UnsupervisedIE} clustering model gets mapped to any intended slot, and how does the evaluation work.

We consider the gold slots as the relation types or role of entities which are annotated inside the documents. For instance, in the context passage \enquote{\textit{Glutamate stimulates glutamate receptor interacting protein 1 degradation by ubiquitin-proteasome system to regulate surface expression of GluR2. Down-regulation of GRIP1 by glutamate was blocked by carbobenzoxyl-leucinyl-leucinyl-leucinal (MG132), a proteasome inhibitor and by expression of K48R-ubiquitin, a dominant negative form of ubiquitin. Our results suggest that glutamate induces GRIP1 degradation by proteasome through an NMDA receptor-Ca2+ pathway and that GRIP1 degradation may play an important role in regulating GluR2 surface expression.}}, the gold tuples annotated are: ``Glutamate [SEP] downregulator GRIP1 and glutamate receptor interacting protein 1". Here, the gold slot is downregulator and the entities involved in this slot are Glutamate, GRIP1 and glutamate receptor interacting protein 1. After running \textit{UnsupervisedIE} initially, we obtain a question-answer pair such as \textbf{Question:} \textit{\enquote{which substance was regulated by glutamate and hence blocked by carbobenzoxyl-leucinyl-leucinyl-leucinal (MG132)?}} - \textbf{Answer:} GRIP1. 

For mapping this predicted question-answer pair to an intended slot, we ask ChatGPT to map the question answer intent to one of the slots and provide the description of each slot: \textbf{``cause"}:``mention of something like what drugs cause which disease", \textbf{``downregulator"}: ``decrease or inhibition effects of any biomedical drug on enzymes or other biomedical species",\textbf{``upregulator"}: ``increase or rise in the effects of any biomedical drug on enzymes or other biomedical species", \textbf{``interacts with"}: ``mention of any adverse effect when two or more biomedical species act together" and  \textbf{``regulator":} ``when there is any binding effect between biomedical species". After the mapping, we use fuzzy matching to determine if the involved entities in the gold tuple \textbf{Glutamate}, \textbf{GRIP1} and \textbf{glutamate receptor interacting protein 1} are present in the predicted QA pair. If yes, we consider that as a true positive and evaluate with respect to gold standard slots using the standard metrics of Precision, Recall and F1-measures.
Moreover, we also merge the results of two or more clusters if those are mapped to similar gold slot, and then again evaluate with the similar metrics.

\begin{figure*}[!t]
\fbox{\includegraphics[width=0.96\textwidth]{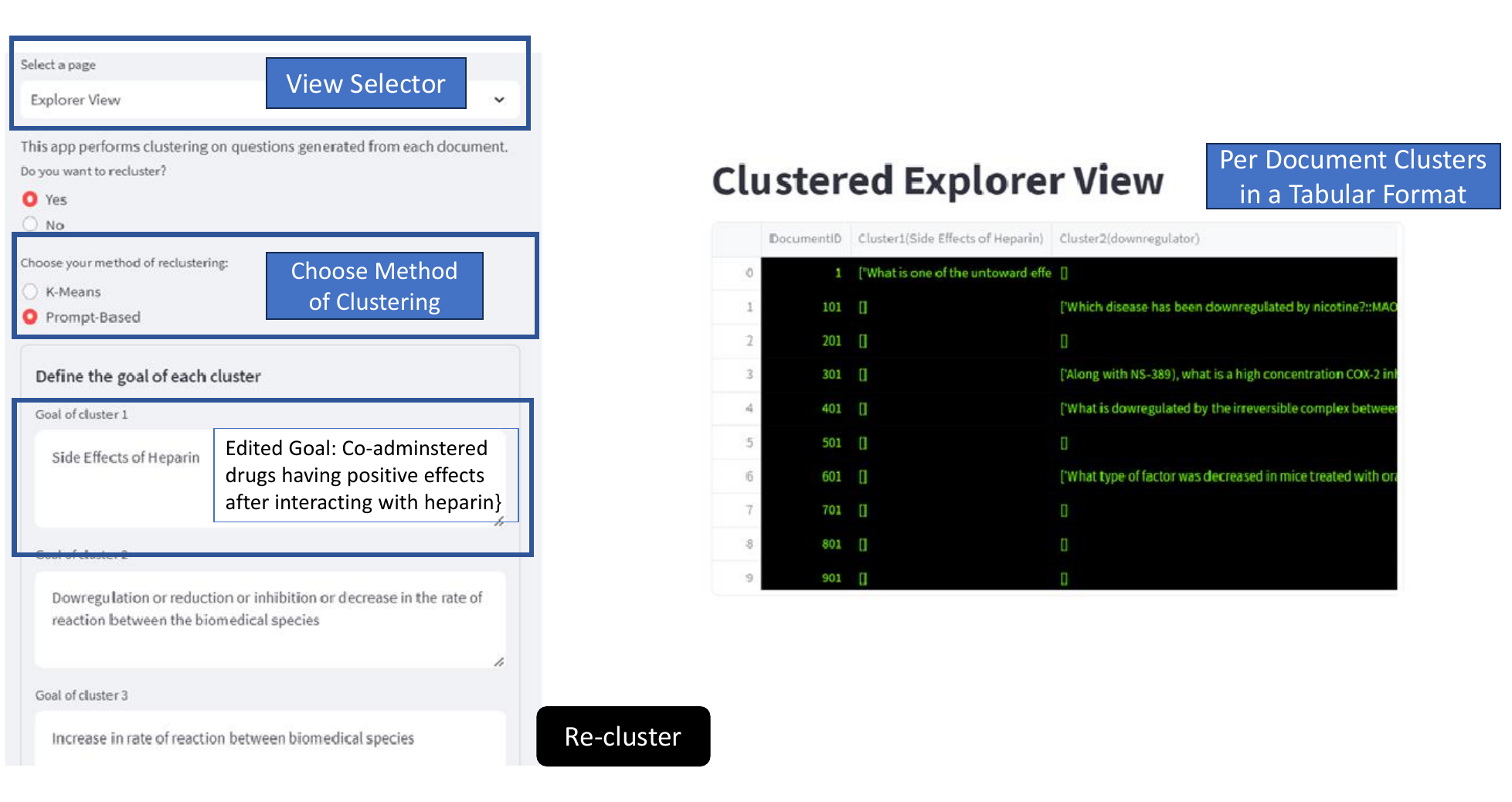}}
\caption{shows the \textit{Explorer View} of \textit{InteractiveIE}. Users can see the clusters generated by the model with the rationales. Based on needs, they can edit the existing goal of \textit{``Side Effects of heparin"} to \textit{``Co-administered drugs having positive effects after interacting with heparin"}. Then users can find the new set of clusters by pressing ``Recluster Button". Based on the goals, the clusters have been named to some slot such as the goal of cluster 3 \textit{``Increase in rate of reaction between biomedical species"} to \textit{``Upregulation"} as seen in Clustered Explorer View. }
\label{fig:explorerview}
\end{figure*}

\begin{figure*}
\fbox{\includegraphics[width=0.98\textwidth]{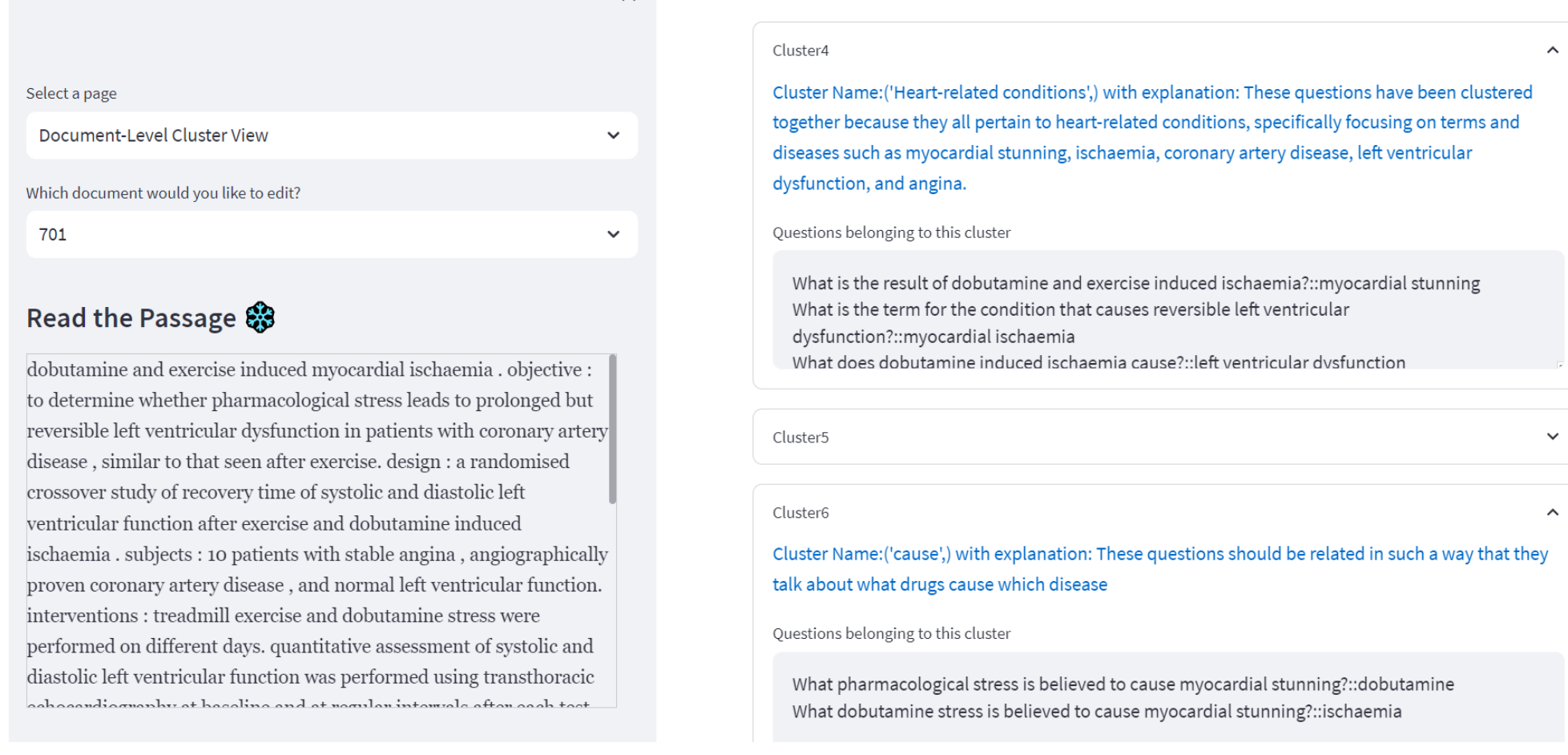}}
\caption{shows the Document-Level Cluster view where the document along with the model-generated clusters are being displayed, users can rearrange, add new or delete existing questions and press ``Refine and Lock" Button to save new explanations of clusters after their edits.}
\label{fig:frameworkoverview}
\end{figure*}

\section{InteractiveIE Framework Overview}
Table~\ref{tab:badclusters-K-means} and~\ref{tab:badclusters-LLM} show two initial clusters generated by \textit{Unsupervised-IE} using $K$-means and LLM-guided clustering respectively.
In this section, we expand on the idea of whether a human-in-the-loop approach can quickly refine these clusters and extract information based on their needs using our \textit{InteractiveIE} interface which comprises of three components:

\paragraph{1. Preprocessing View}
\label{preprocessingview}
This component (Figure \ref{fig:preprocessing}) helps to select documents from a pool of documents and then generates clusters based on an \textit{Unsupervised-IE} pipeline which can be illustrated as:  1) using named entity recognition tool for finding the potential entities from documents (spacy-based NER models: en-core-web-sm, en-core-web-md, en-ner-bc5cdr-md and en-core-sci-scibert), 2) followed by automatically generating factoid questions conditioned on the document and the tagged entity mentions ~\citep{lewis-etal-2020-bart,lewis-etal-2021-paq} (T5~\footnote{\url{https://huggingface.co/iarfmoose/t5-base-question-generator}} or BART\footnote{\url{https://huggingface.co/tszocinski/bart-base-squad-question-generation}}) (Steps 1 and 2 of \ref{fig:frameworkoverview}), 3) embedding those (TF-IDF from Scikit-learn or SentenceTransformers\footnote{\url{https://huggingface.co/sentence-transformers}}) and 4) grouping those ($K$-means or ChatGPT) into user-specified number of clusters.

\begin{table}[!t]
    \small
    \centering
    \scalebox{1}{
    \begin{tabular}{p{0.07\textwidth}|p{0.37\textwidth}}
    \toprule
   \textbf{Clusters} & \textbf{Questions} \\
    \toprule
     Adverse Effects & What disease is adversely caused due to the intake of heparin?  \\
      \midrule
     Drug Effects & what drug may inhibit the metabolism of mifepristone?, What are the side effects of heparin?, What can be the subacute effects of cocaine?, What causes heparin? \\
    \bottomrule
    \end{tabular}
    }
    \caption{shows the output from \textit{Unsupervised IE} pipeline using $K$-means clustering. The clusters do not look semantically coherent since \textit{``What causes heparin?''} is not a great example of the drug effects cluster. 
    }
    \label{tab:badclusters-K-means}
\end{table}

\begin{table}[!t]
    \small
    \centering
    \scalebox{1}{
    \begin{tabular}{p{0.07\textwidth}|p{0.37\textwidth}}
    \toprule
   \textbf{Clusters} & \textbf{Questions} \\
    \toprule
     Heparin Side Effects & What are the side effects of heparin?, What disease is adversely caused due to the intake of heparin?, What causes heparin?  \\
      \midrule
     Drug Effects & what drug may inhibit the metabolism of mifepristone?, What can be the subacute effects of cocaine?\\
    \bottomrule
    \end{tabular}
    }
    \caption{shows the output from \textit{Unsupervised IE} pipeline using ChatGPT-based clustering. The clusters look more semantically coherent in this case compared to the ones generated by K-Means clustering, but the cluster names are different in this case.
    }
    \label{tab:badclusters-LLM}
\end{table}

\paragraph{2. Clustering Explorer View}
\label{explorerview}
After landing on this page (Figure~\ref{fig:explorerview}), users will be shown a dataframe with columns as cluster names and rows as selected documents where they can explore the machine generated questions grouped together in various clusters corresponding to each of their selected documents along with the explanations of why it has been clustered in a specific way (Step 3, 4 in Figure \ref{fig:humanaidiagram}).
Then we further allow the users to refine the rationales behind each cluster according to their need. 
We allow two types of clustering: 
\paragraph{A. $K$-means based:} In this case, the users can tweak the mean (representative questions) of each cluster based on the kind of information they are interested to extract in that cluster. For instance, from Tables~\ref{tab:badclusters-K-means} and~\ref{tab:badclusters-LLM} if the user is more interested in getting two clusters having intents as \textit{\enquote{Drug Side Effects}}, \textit{\enquote{Downregulation}} then he will assign representative questions in such a way that it gets reflected after reclustering. They might probably assign \textit{\enquote{What are the side effects of heparin}} in one cluster and \textit{\enquote{what drug may inhibit the metabolism of
mifepristone?}} in another cluster as representative questions.
After the humans edit the goal of each cluster, they can press the \textbf{\enquote{Recluster}} button. After this, the $K$-means algorithm reclusters other questions by calculating pairwise similarity between the representative questions and other questions and assigns to the cluster which has the most similar mean (Step 5, 6 in Figure \ref{fig:humanaidiagram}). 
\paragraph{B. LLM-guided natural description based:} Here the humans will be initially shown the model generated explanations for each clusters, where they can tweak the prompt/natural description in order to specify what is the goal or intent of putting questions in that cluster. For instance: in the Table \ref{tab:badclusters-LLM}, the user edits the description from \textit{\enquote{Heparin Side Effects}} to \textit{\enquote{Side Effects of various drugs}} and \textit{\enquote{Drug Effects}} to \textit{\enquote{Decrease in rate of reaction of biomedical species}}. Now, on pressing the \textbf{\enquote{Recluster}} button, human-edited goals are used to further recluster questions using LLM and rendered back for exploration (Step 5, 6 in Figure \ref{fig:humanaidiagram}) [Final clusters in Table \ref{tab:finalLLM}].

\paragraph{3. Document-Level Cluster View}
\label{doccluster}
After looking at the clusters in Explorer View (\ref{fig:explorerview}), the humans might want to modify questions in each cluster for each document in the Document-Level Cluster View (Figure \ref{doccluster}) because the automatically generated questions might be ill-formed, or unnecessarily placed in some cluster where the user does not want those to be, some might be redundant too.
So, this view lets the users read the documents and decide which operations to perform in order to modify the induced clusters in each document:
\paragraph{A. Rearrange Questions:}
If a question looks unfitting in a cluster and the user feels it better fits another cluster, they can rearrange the questions from unfitting to fitting cluster. In Table~\ref{tab:badclusters-LLM}, the question \enquote{\textit{What can be the subacute effects of cocaine?}} can be moved to the cluster fitting the information need of side effects of any drugs.

\paragraph{B. Delete questions:} If the question looks unfitting to none of the clusters, the user can delete that information altogether. For instance, \enquote{\textit{What causes heparin?}} is unnecessary and can be deleted.
\paragraph{C. Ask new questions:} If the user feels that they need to ask more questions or even edit an existing question, they can do that as well. We use distillbert-based extractive QA model \cite{distillbert} to extract answers for new questions.

After the user performs these operations, they can press \textbf{``Infer Explanations"} button to see what did their actions of modifying the clusters result in the overall explanation of the cluster content. 
Finally, the new explanations are shown to the users to understand how much their actions have modified the goal of the clusters, and whether they want to edit that further (Refer to \ref{appendix:inference} for details).

\paragraph{Dynamic Slot Mapping and Evaluation}
\label{appendix:dynamic}
Since the users are provided with the agency to change the goal of each cluster, they can edit and refine that through the \textit{``Clustering Explorer View"}. After the goals are refined, we again prompt GPT3.5 (ChatGPT) to recluster based on the new user-specified goals. The prompt chosen is \textit{\enquote{Can you recluster these questions '+allquestions+' into '+K+ ' clusters, among which each cluster  should contain information about '+(listofrationales) + ".A single question should not be placed in more than one cluster. Format your response as a list of JSON Objects where the keys can be ``ClusterID", ``ClusterName" and ``ClusterContent" where clusterid should start as Cluster1.}}

Even when the cluster contents in \textit{``Document-Level Cluster View"} changes, the clusters ar generally updated  and we call the similar prompt to map the clusters containing generated QA pairs to a short description. 
If the description fits to one of the gold intents, then we evaluate based on the fuzzy matching approach as discussed in~\ref{par:evaluation}, if the description doesn't match, then we refrain from evaluating with the gold standard slots.

\section{Implementation Details}
\textit{InteractiveIE} has been developed using Streamlit\footnote{\url{https://streamlit.io/}} with programming language as python and Sqlite3\footnote{\url{https://www.sqlite.org/index.html}} as database. We make use of OpenAI's \textit{gpt3.5-turbo} for running our LLM-based clustering.

\begin{table}[!t]
    \small
    \centering
    \scalebox{1}{
    \begin{tabular}{p{0.12\textwidth}|p{0.32\textwidth}}
    \toprule
   \textbf{Clusters} & \textbf{Questions} \\
    \toprule
     Side Effects of any drug & What are the side effects of heparin?, What disease is adversely caused due to the intake of heparin?, What causes heparin?,  What can be the subacute effects of cocaine? \\
      \midrule
     Decrease in rate of
reaction of biomedical species & what drug may inhibit the metabolism of mifepristone?\\
    \bottomrule
    \end{tabular}
    }
    \caption{The human edits the goal of cluster as ``Decrease in rate of reaction of biomedical species" which inturn gets mapped to the slot: Downregulation. The table shows the output from \textit{InteractiveIE} pipeline after human-edits in Explorer View, where the downregulator slot is only represented by the question \textit{what drug may inhibit the metabolism of mifepristone?}
    }
    \label{tab:finalLLM}
\end{table}

\begin{table*}
\small
\setlength\tabcolsep{5pt}
\centering
\begin{tabular}{ l c c c c  c} 
\toprule
&  Upregulation & Downregulation & Interacts with & Cause & Regulation \\
\midrule
Random & 0.05 & 0.17 & 0.03 & 0.15 & 0.07 \\
\cite{angeli-etal-2015-leveraging} & 0.17 & 0.15 & 0.24 & 0.23 &  0.18 \\
\cite{clausie} & 0.11 & 0.21 & 0.28 & 0.24 &  0.16 \\
\cite{promptore} & 0.34 & 0.41 & 0.51 & 0.48 &  0.39 \\
\cite{zhang-etal-2023-human} & 0.22 & 0.36 & 0.31 & 0.50 &  0.40 \\
\midrule
UnsupIE (KMeans) & 0.41 & 0.43 & 0.52 &  0.53 & 0.44 \\
UnsupIE (LLM) & 0.45 & 0.47 & 0.52 &  0.53 & 0.47 \\
 \bottomrule
\end{tabular}
\caption{
F1-scores of slots mapped by our \textit{UnsupervisedIE} baselines (using both Kmeans and LLM) compared to the existing unsupervised Information Extraction models on Biomedical Slot Dataset. OpenIE and ClauseIE also have relatively low F1 scores around 0.11-0.28, underperforming PromptORE. The proposed UnsupervisedIE methods using KMeans and LLMs outperform PromptORE, achieving the best results with F1 scores of 0.41-0.53.
}
\label{tab:unsupbiomedical}
\end{table*}

\begin{table*}
\small
\setlength\tabcolsep{5pt}
\centering
\begin{tabular}{ l c c c c  c} 
\toprule
&  Agreement Date & Effective Date & Expiry Date & Termination Date & Contract Name \\
\midrule
Random & 0.16 & 0.27 & 0.10 & 0.01 & 0.05 \\
\cite{angeli-etal-2015-leveraging} & 0.11 & 0.13 & 0.12 & 0.13 &  0.08 \\
\cite{clausie} & 0.20 & 0.20 & 0.16 & 0.20 &  0.26 \\
\cite{promptore} & 0.37 & 0.56 & 0.24 & 0.27 &  0.44 \\
\cite{zhang-etal-2023-human} & 0.34 & 0.45 & 0.32 & 0.28 &  0.48 \\
\midrule
UnsupIE (KMeans) & 0.34 & 0.55 & 0.27 &  0.29 & 0.48 \\
UnsupIE (LLM) & 0.37 & 0.58 & 0.28 &  0.31 & 0.50 \\
 \bottomrule
\end{tabular}
\caption{
F1-scores of slots mapped by our \textit{UnsupervisedIE} baselines (using both Kmeans and LLM) compared to the existing unsupervised Information Extraction models on CUAD Dataset. The proposed UnsupervisedIE methods using KMeans and LLMs significantly outperform the baselines, with F1 scores of 0.27-0.58. In particular, the LLM-based clustering achieves the best F1 scores for 4 out of 5 slots, showing the effectiveness of this approach.
}
\label{tab:unsuplegal}
\end{table*}

\begin{figure*}[!t]
    \centering
\begin{subfigure}[t]{.40\columnwidth}
    \centering
    \includegraphics[width=\columnwidth]{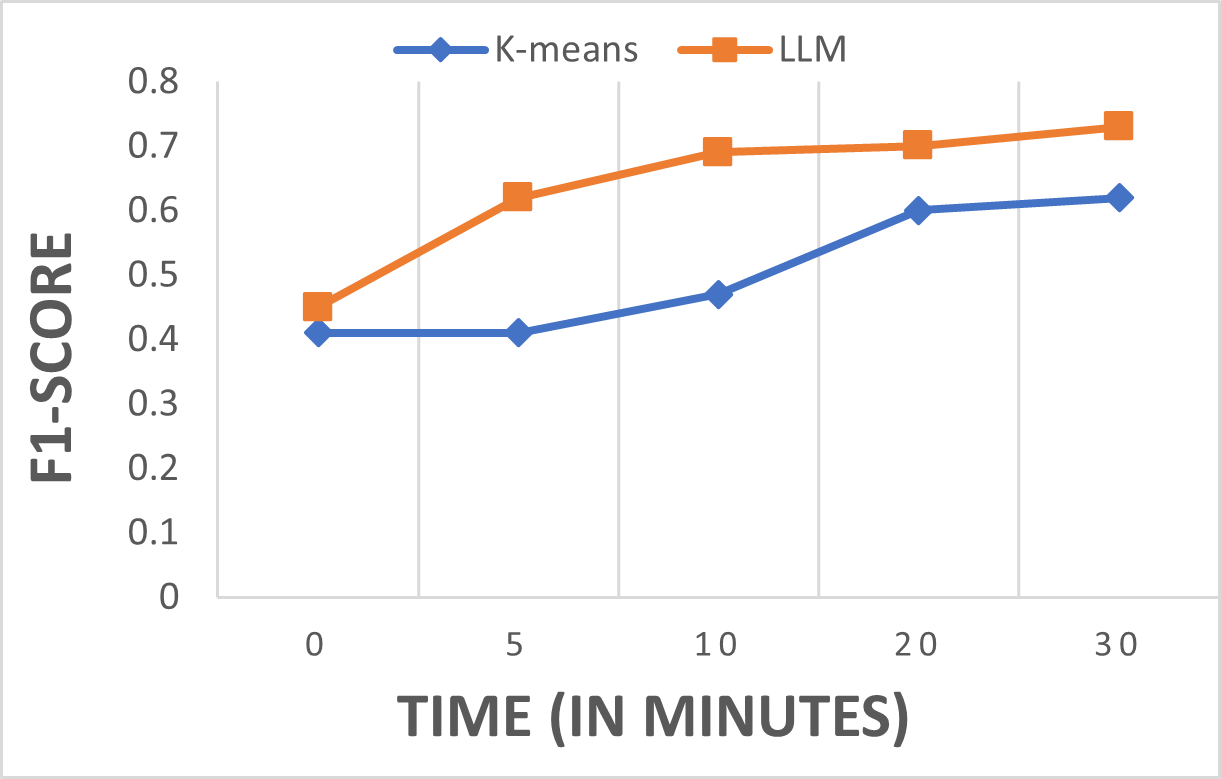}
    \caption{Upregulation}
\end{subfigure}
\begin{subfigure}[t]{.40\columnwidth}
    \centering
   \includegraphics[width=\columnwidth]{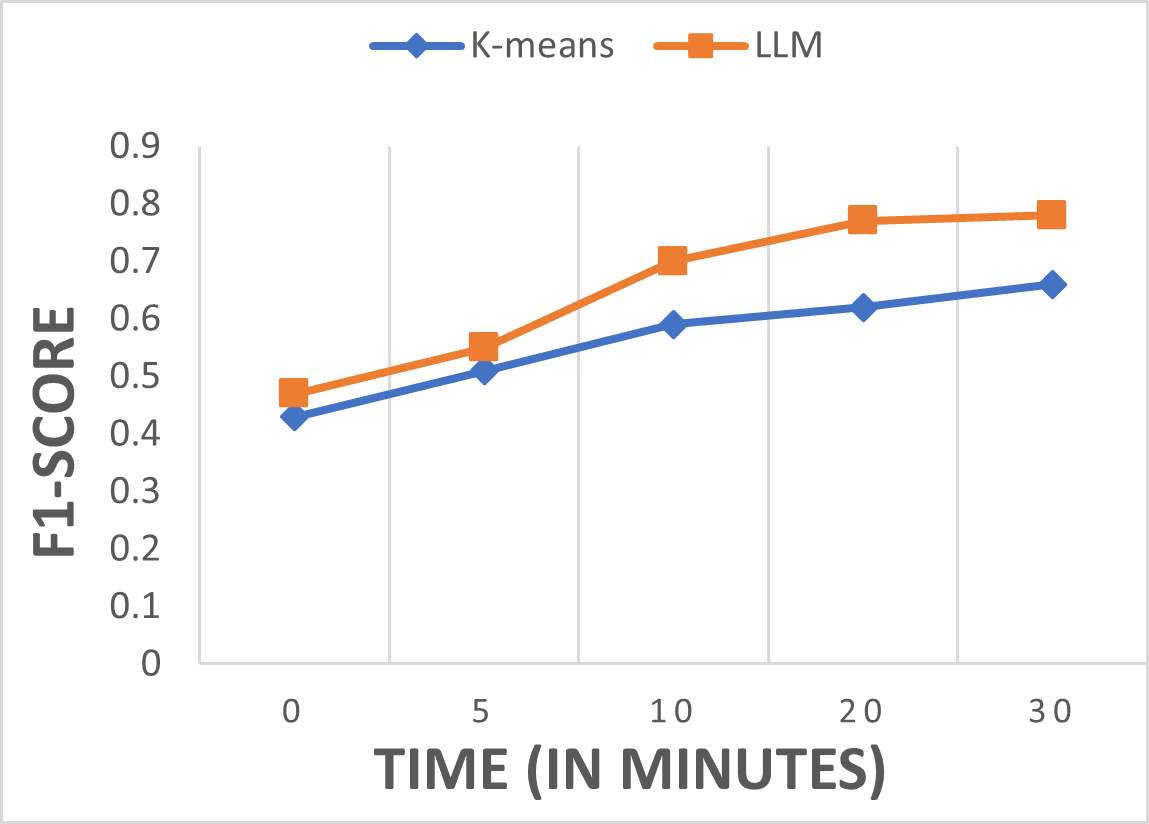}
    \caption{Dowregulation}
\end{subfigure}
\begin{subfigure}[t]{.40\columnwidth}
    \centering
    \includegraphics[width=\columnwidth]{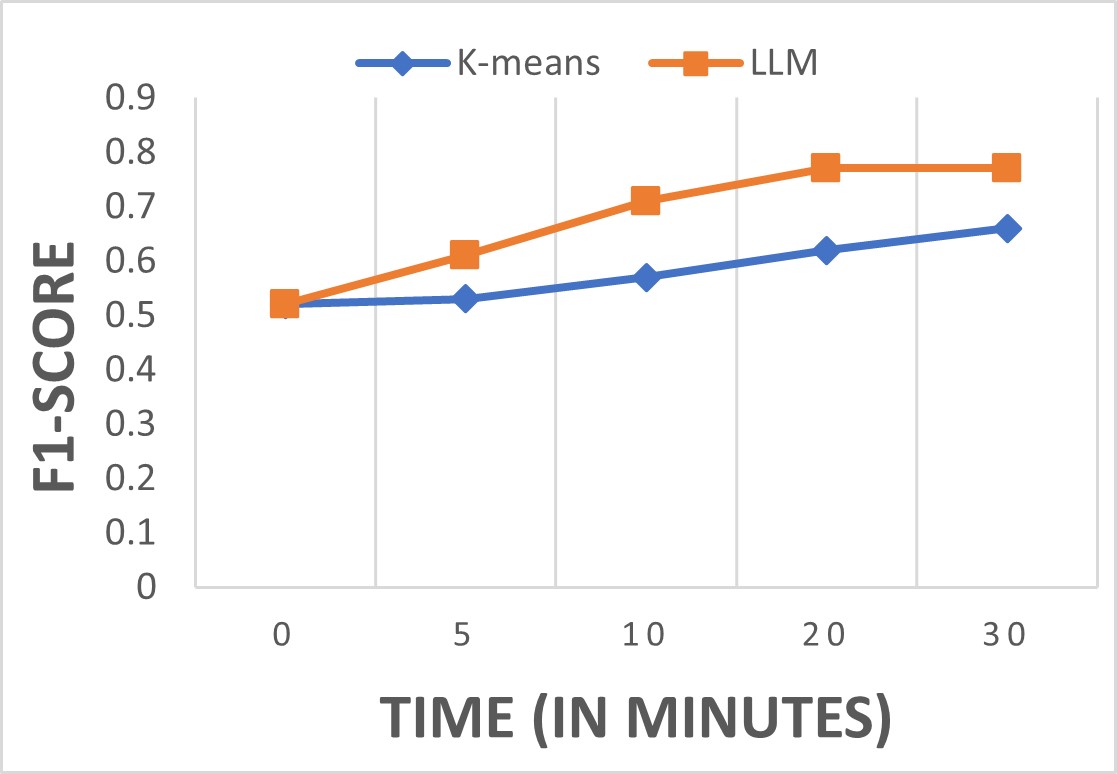}
    \caption{Interacts with}
\end{subfigure}
\begin{subfigure}[t]{.40\columnwidth}
    \centering
    \includegraphics[width=\columnwidth]{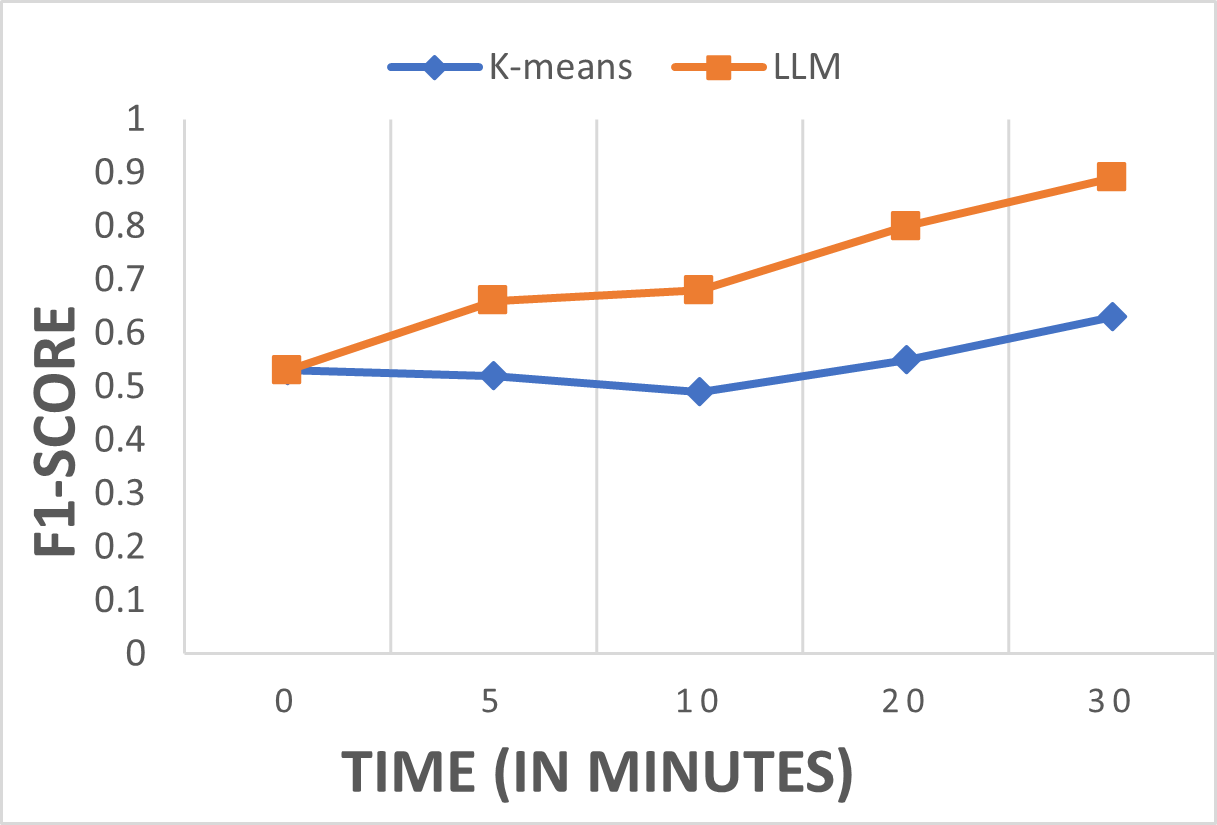}
    \caption{Cause}
\end{subfigure}
\begin{subfigure}[t]{.40\columnwidth}
    \centering
    \includegraphics[width=\columnwidth]{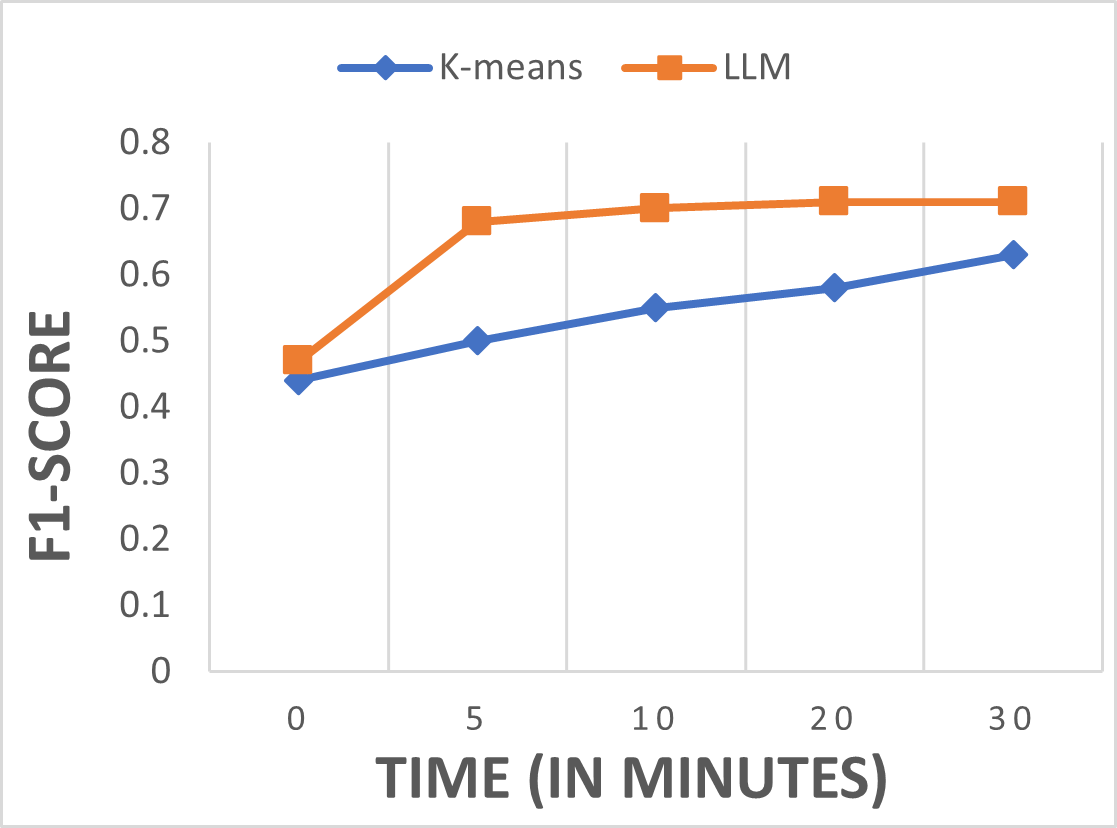}
    \caption{Regulator}
\end{subfigure}
\caption{Average F1-scores of 3 users at different time stamps for each slot in Biomedical Dataset. At time 0, \textit{UnsupervisedIE} clusters are shown initially and the participants kept interacting with \textit{InteractiveIE} for 30 minutes.}
\label{fig:biohumanperf}
\end{figure*}

\begin{figure*}[!t]
    \centering
\begin{subfigure}[t]{.40\columnwidth}
    \centering
    \includegraphics[width=\columnwidth]{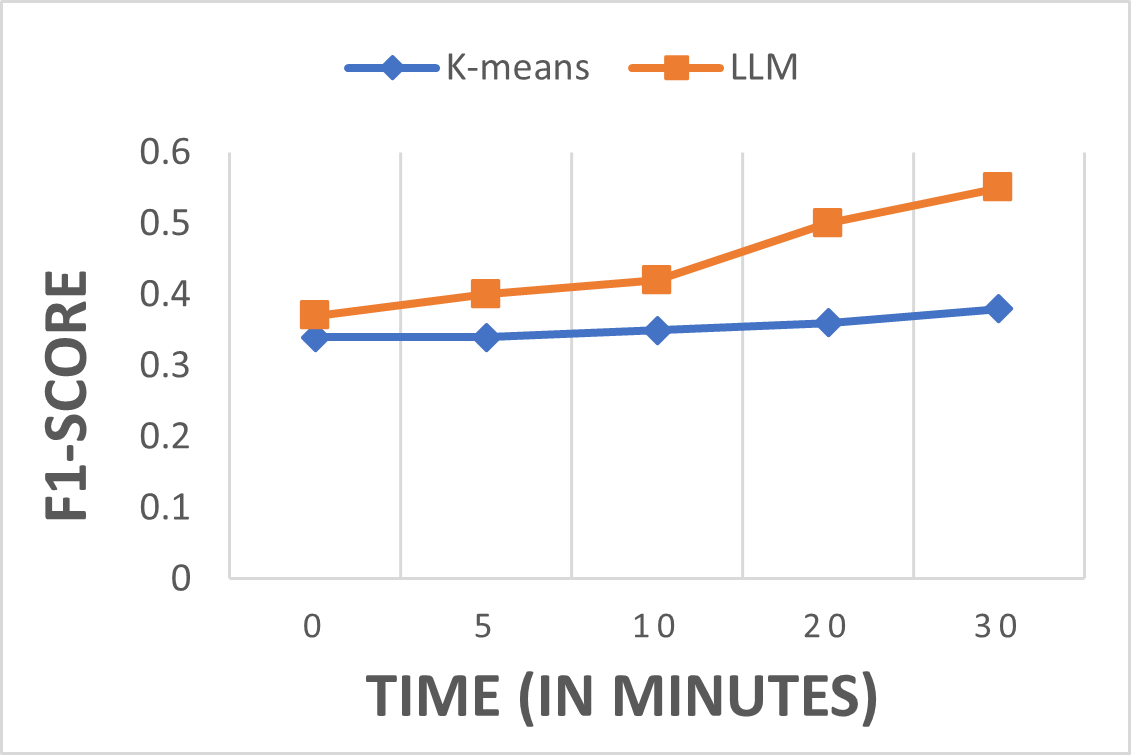}
    \caption{Agreement Date}
\end{subfigure}
\begin{subfigure}[t]{.40\columnwidth}
    \centering
    \includegraphics[width=\columnwidth]{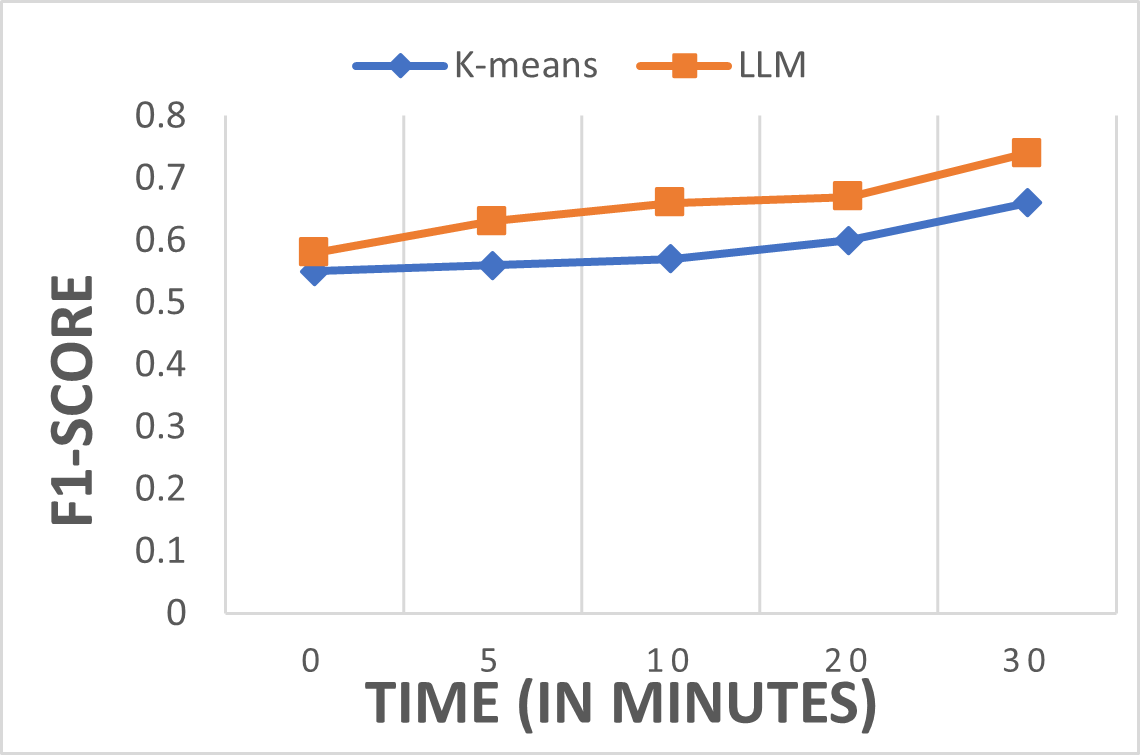}
    \caption{Effective Date}
\end{subfigure}
\begin{subfigure}[t]{.40\columnwidth}
    \centering
    \includegraphics[width=\columnwidth]{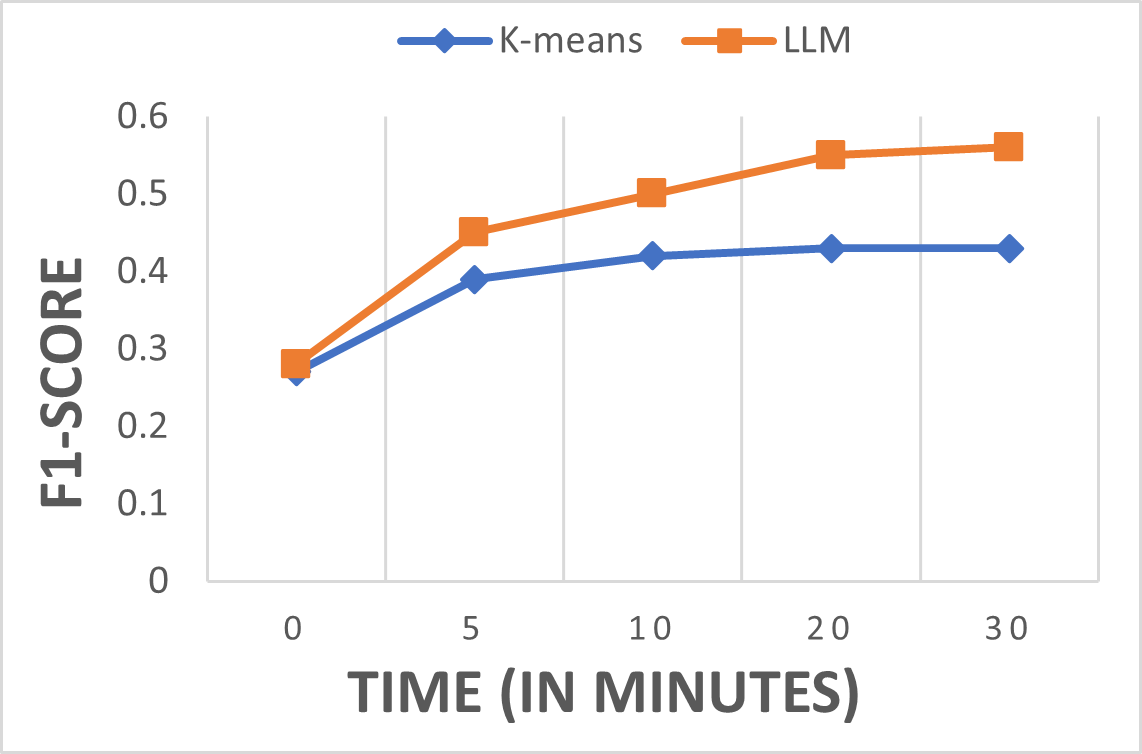}
    \caption{Expiry Date}
\end{subfigure}
\begin{subfigure}[t]{.40\columnwidth}
    \centering
    \includegraphics[width=\columnwidth]{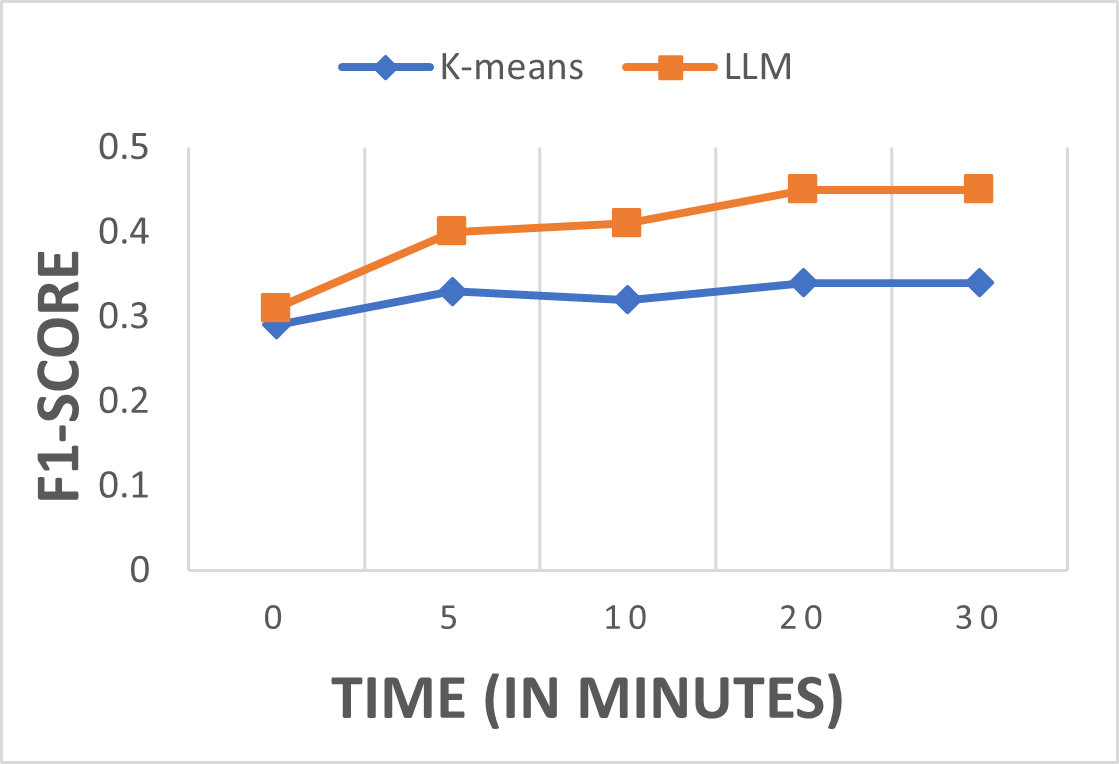}
    \caption{Termination Date}
\end{subfigure}
\begin{subfigure}[t]{.40\columnwidth}
    \centering
    \includegraphics[width=\columnwidth]{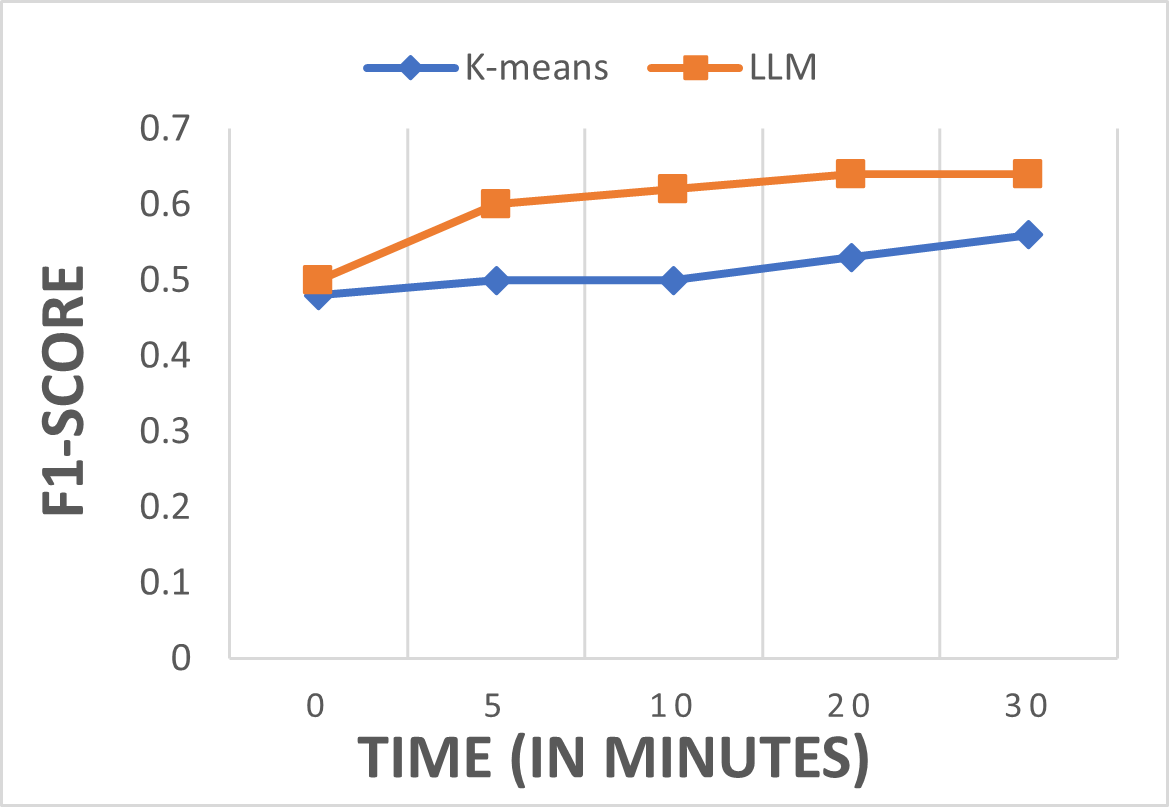}
    \caption{Contract Name}
\end{subfigure}
\caption{Average F1-scores of 3 users at different time stamps for each slot in CUAD Dataset. At time 0, \textit{UnsupervisedIE} clusters are shown initially and the participants kept interacting with \textit{InteractiveIE} for 30 minutes.}
\label{fig:cuadhumanperf}
\end{figure*}

\section{Experiments and Evaluation}

\subsection{How good is the UnsupervisedIE pipeline for IE schema induction?}
We evaluate the effectiveness of our question-driven and clustering based UnsupervisedIE pipeline over existing methods. 
We extract triples from documents and map each triple to the nearest slot (as explained in Datasets) based on fuzzy matching technique using 
some existing state-of-the-art unsupervisedIE baselines such as: 1) \textbf{OpenIE} \cite{angeli-etal-2015-leveraging}, 2) \textbf{ClauseIE} \cite{clausie} 3) \textbf{PromptORE} \cite{promptore} which extracts some of the trigger words surrounding the context, followed by clustering and slot mapping. However, our methods do not rely on heuristics to find trigger words between two or more entities in the sentences, instead consider the overall context to ask questions conditioned on the tagged entities.

Besides, we also compare our approach with the existing GPT-3 generated Subject-Verb-Object (SVO) tuple generation techniques as proposed in \cite{zhang-etal-2023-human}. Here, we extract the entities from the corpus and then use the same few-shot prompt (both instruction and demonstrative examples) to extract SVO tuples.

We run the existing unsupervised baselines along with our proposed \textit{UnsupervisedIE} approach on the choosen documents from both the two domain-specific datasets.
Tables~\ref{tab:unsupbiomedical} and~\ref{tab:unsuplegal} show the F1-scores of each slot type for both biomedical and legal datasets.
From the results, we can infer that \textbf{PromptORE} is the strongest existing baseline among all in terms of average F1-scores (0.43 on biomedical and 0.38 on legal). 
Our \textit{UnsupervisedIE} using Kmeans (0.46 on biomedical and 0.38 on legal) and LLMs (0.48 on biomedical and 0.41 on legal) exhibit 
higher performance compared to the state-of-the-art existing \textbf{PromptORE} baselines. Overall, we infer that clustering of trigger words/questions has proved to be a better method compared to triple based methods (like OpenIE and ClausIE). 

\subsection{Can humans improve schema induction with minimal supervision?}
We evaluate the effectiveness of our \textit{InteractiveIE} interface against \textit{UnsupervisedIE} approaches of extracting information on-the-fly by carrying out human studies on two domain-specific datasets.

\paragraph{Datasets} 60 documents were sampled from two datasets (30 from each\footnote{These were selected to represent a diverse set of
slots.}): 1) Biomedical Slot Filling \cite{papanikolaou-etal-2022-slot} dataset, where each instance contains Subject, Relation, Object (SVO) triple and the text where it was found, so we can reuse for slot filling. We consider  five types of slots here \enquote{\textit{Upregulator}}, \enquote{\textit{Downregulator}}  , \enquote{\textit{Cause}}, \enquote{\textit{Interacts with}} and \enquote{\textit{Regulator}}. 
In the CUAD dataset, all the factoid questions are converted into slot-filler templates to make this dataset suitable for our slot-filling task. We mainly consider those question-answer pairs which are extractive spans in the document: \enquote{\textit{On what date is the contract is effective?}} \textbf{(Effective Date)},  \enquote{\textit{On what date will the contract’s initial term expire?}} \textbf{(Expiry Date)}, \enquote{\textit{What is the date of the contract?}}  \textbf{(Agreement Date)}, \enquote{\textit{What is the notice period required to terminate renewal?}} \textbf{(Termination Date)}, \enquote{\textit{What is the name of the contract?}} \textbf{(Contract Name)}

\paragraph{Human Study Recruitment}
\label{humanstudy}
Our user study was not limited to the individuals who are well-versed in the concepts of Machine Learning or Natural Language Processing, we wanted to verify if the participants can understand what does a semantically coherent cluster look like. For this, we recruited those participants with their native language as English. Out of three, only one participant had prior experience on NLP. In order to familiarize them with the clustering task, we asked them to solve a simple assignment as described in figure~\ref{fig:assignment}. We recruited those participants who could successfully complete the task without any difficulty. 
Prior to the study, we collected consent forms for the workers to agree that their answers would be used for academic purposes. All the involved participants gave their consent to disclose their interactions with the interface.

\paragraph{Participants and Evaluation} A total of three participants joined the experiment. All the participants were not previously exposed to this task and interface. To help them become familiar, they were first asked to read 50 questions, answers and mapped slots for 5 intents for both the domains. Then, the participants were asked to interact with the interface for 1 hour (30 minutes for each domain) during which we record their interactions.
After we run the \textit{UnsupervisedIE} pipeline on the selected documents, the participants were asked to extract information based on the above-mentioned intents from these documents, without revealing true answers. 
Their goal was to maximize information extracted in both domains.
We evaluate the slot mapping performance in both \textit{UnsupervisedIE} and \textit{InteractiveIE} using ChatGPT-based dynamic slot mapping followed by fuzzy-matching algorithm of question-answer pairs to the desired slots using Precision, Recall and F1-scores (See \ref{appendix:dynamic}).

\paragraph{Evaluation through Human Study}
We compare our semi-supervised \textit{InteractiveIE} against \textit{UnsupervisedIE} by evaluating the change in the slot mapping accuracy over time. 
A generic observarion in figure \ref{fig:biohumanperf} and figure \ref{fig:cuadhumanperf} is that the humans could achieve higher F1-scores compared to \textit{InteractiveIE} on slot mapping within 30 minutes.
We conduct deeper analysis of performance improvement based on three research questions: a) which user actions cause performance improvement? b) which clustering algorithm is effective in terms of performance improvement?, c) how useful are the clustering model explanations in guiding the users to extract slots correctly? We discuss our findings below:
\paragraph{A) LLM clustering improves slot mapping with lesser number of user actions than K-means.}  Overall, LLMs outperform K-Means by an average of 11.4\% on legal and 13.1\% on biomedical documents. From the figures \ref{fig:biohumanperf} and \ref{fig:cuadhumanperf}, we observe that in first 10 minutes, LLMs improve by 11.4\% compared to 5.6\% in last 20 minutes, whereas K-Means improve by 4.6\% compared to 4.2\% in last 20 minutes. This is because in first 10 minutes, users edit overall goals in "Explorer view" and then tweaked the prompts. The latter action leads to lesser improvement rate. Whereas, users took uniform time to tweak mean representative questions (in K-Means) to improve mapping of desired slots.  
\paragraph{B) Reclustering of questions in Document-Cluster View has been more effective compared to other actions.} During the user interactions, we qualitatively investigate that 
in "Document-Level Cluster View". We observe that 76\% of the times, the users re-arranges questions among the clusters to investigate if the slot mapping accuracy improves, and it finally did improve. One of the users' action reported that during reclustering, he edited the question to make it sensible. However, only 1 user tried to ask more questions in the individual documents but that eventually reduced his average F1 of slot mapping by 3\%. 
\paragraph{C) Inference of Explanations after user edits helped to improve performance consistently.} From user actions, we empirically observe that whenever the users pressed the \textbf{"Infer Explanations"} button before saving their changes, 93\% of the times this action led to improvement in slot mapping F1 compared to their performance in previous step. From this, we can infer that clustering model explanations guided the users in this task.

\section{Discussion and Conclusion}
We introduce a human-in-the-loop IE interface powered by clustering and explanation generation capabilities of ChatGPT and K-means algorithms as the backbone.
Since our method is pivoted on factoid question generation (which is a real-time proxy of IE tasks), followed by LLM-guided clustering, this method is scalable to any IE task and domain. We involve humans to judge the grouping of intents to extract information based on their dynamic needs on-the-fly. With empirical evaluations, we show that our system can efficiently extract relevant information based on need.

\section*{Limitations}
We have a few limitations in our approach. First, we have conducted experiments with a small set of users and we plan to scale it up in the future. We will eventually segregate the pool of participants into two groups: participants with domain knowledge and no domain knowledge. This will help us analyze whether domain-specific knowledge is required to extract more useful information from such documents. Second, our experiments are based on two domain-specific datasets, therefore, we hope to experiment on different tasks and datasets where manual data annotation is an expensive affair, such as  non-English datasets (mainly low-resource languages). Finally, some participants wanted to take a look at interactive TSNE plots at each step of their interactions with the interface, particularly when they are tweaking the number of clusters in the preprocessing view. As a next version of the interface, we hope to include both extrinsic and intrinsic evaluation in order to provide better guidance to the users.

\section*{Ethics Statement}
The experiments performed in this study involved
human participants. All the experiments involving human evaluation in this paper were exempt under institutional IRB review. We recruited participants for our human study using Upwork and we have fairly compensated all the Upwork freelancers involved in this study, at an average rate of $15.00$ USD per hour (respecting their suggested Upwork hourly wage).
Prior to the study, the participants provided explicit consent to the participation and to the storage, modification and distribution of the collected data.
All the involved participants gave their consent to disclose their interactions with the interface.
The documents used in the study are distributed under an open license.

\bibliography{interactiveIE}
\bibliographystyle{acl_natbib}

\appendix

\section{Example Appendix}

\begin{figure*}[!t]
\includegraphics[width=\linewidth]{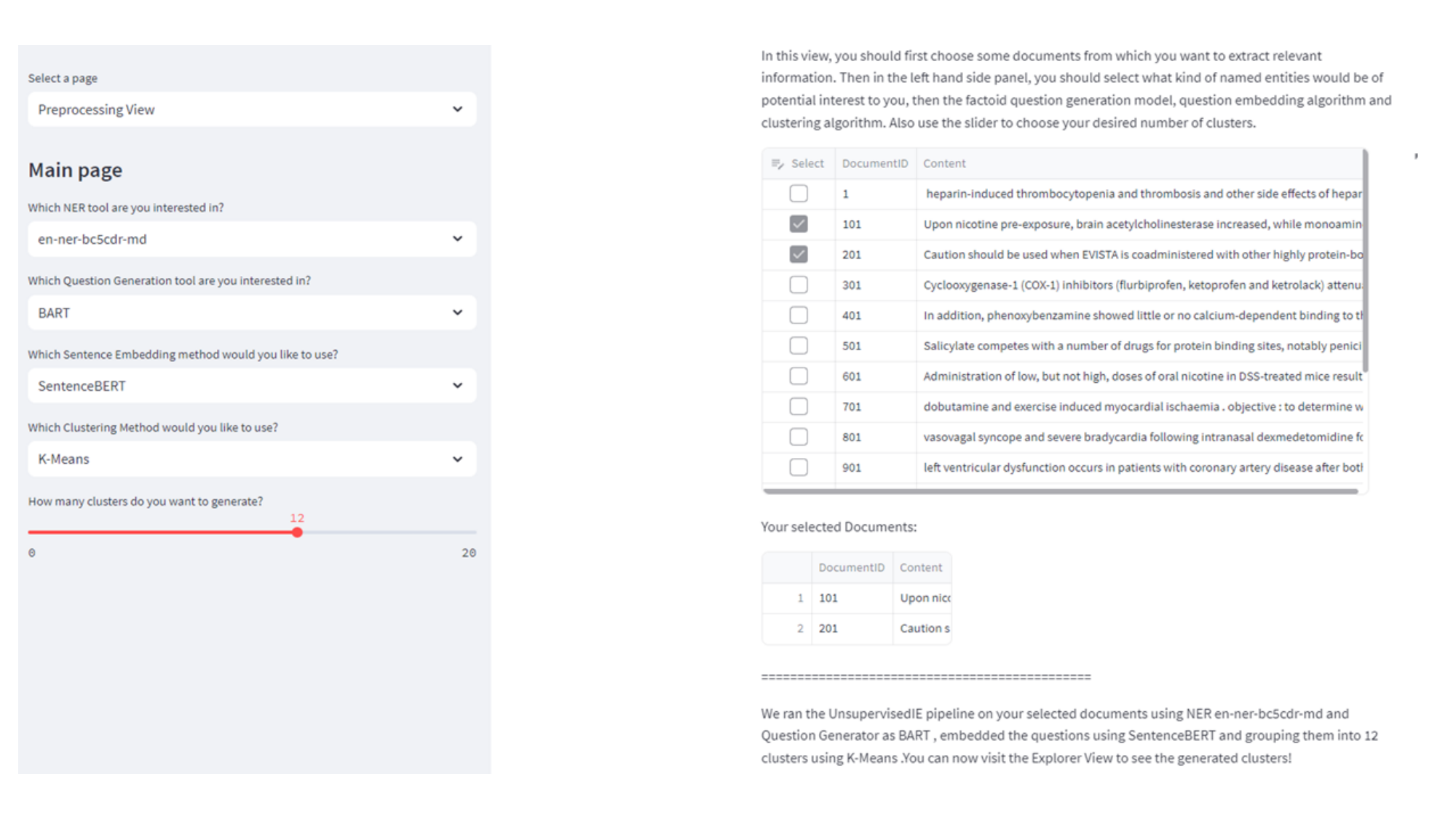}
\caption{First step of running the preprocessing pipeline on the user-specified needs. The user can choose relevant documents, NER model, Question generation model, sentence embedding model, clustering algorithm and number of clusters to group the question-answer pairs into.}
\label{fig:preprocessing}
\end{figure*}

\begin{figure*}[!t]
\includegraphics[width=\linewidth]{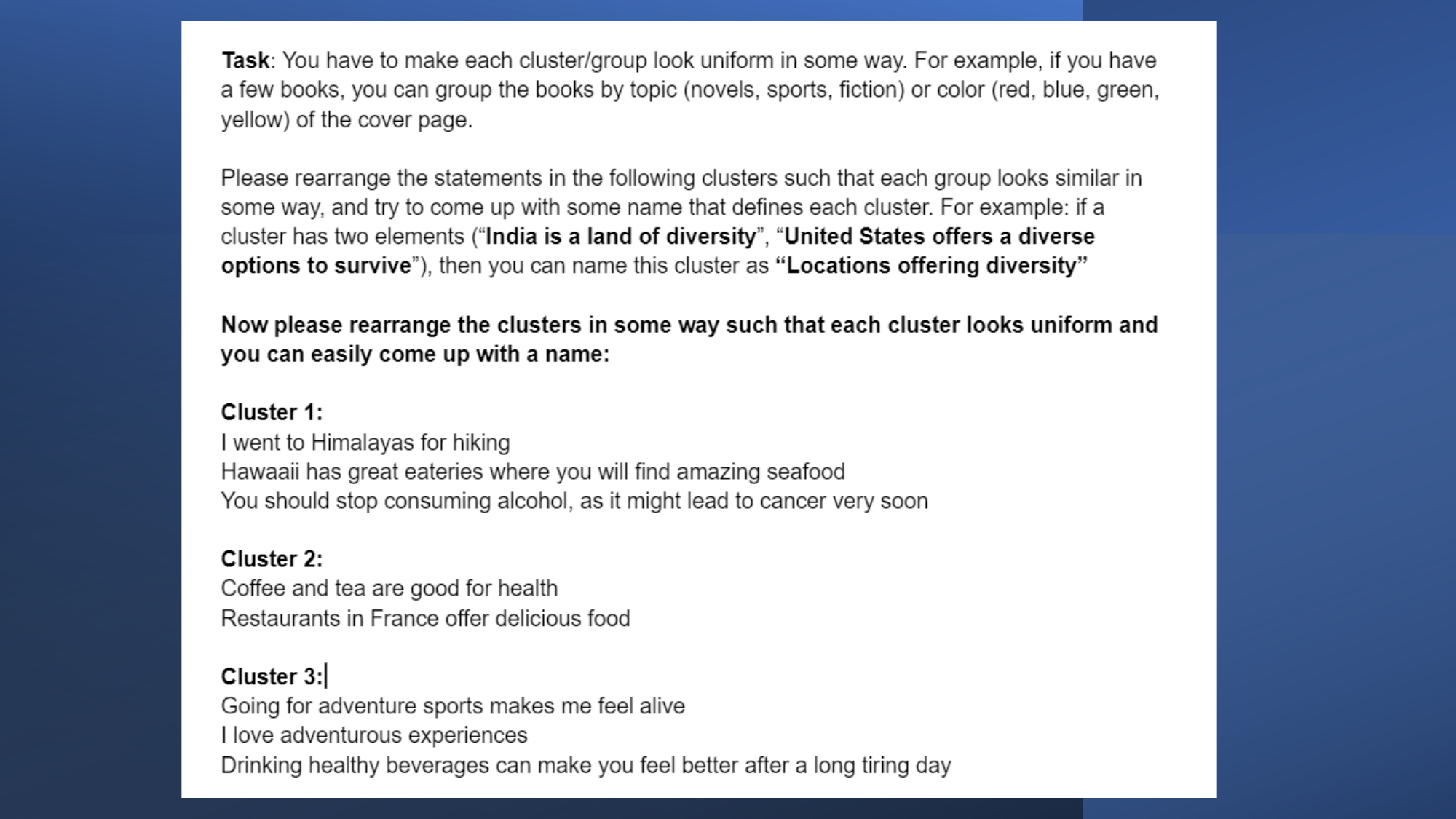}
\caption{Clustering Assignment used for recruiting participants.}
\label{fig:assignment}
\end{figure*}

\label{appendix:evaluation}

\begin{figure*}[!t]
\includegraphics[width=\linewidth]{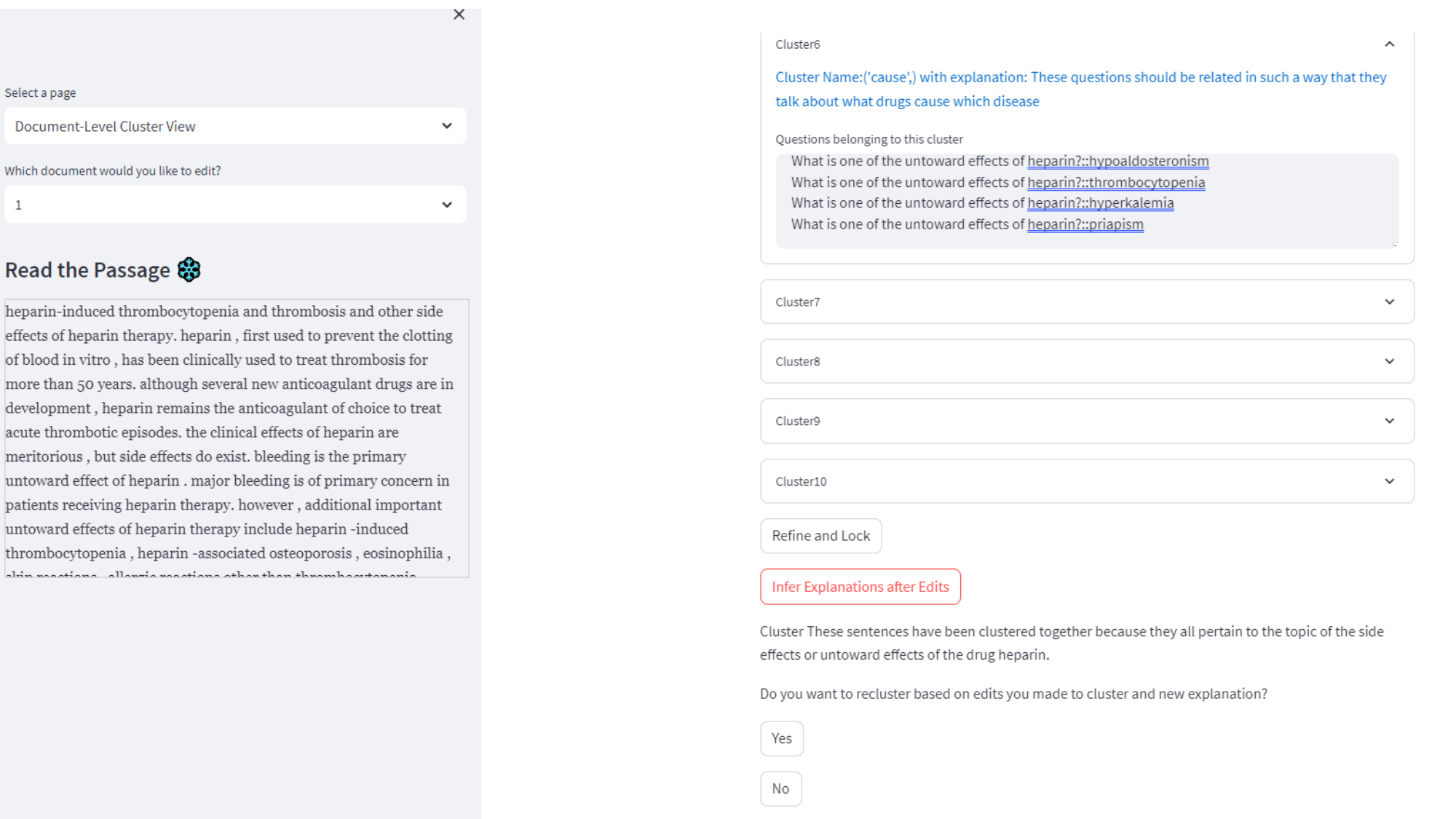}
\caption{Infer Explanations Functionality in the ``Document-Level Cluster view"}
\label{fig:inferexp}
\end{figure*}

\subsection{Prompt-1}
\label{appendix:promptA}
\enquote{\textit{Can you generate an explanation of why these questions: " + (questions)+ "have been clustered together? Or in other words, what makes this cluster semantically coherent? Provide two explanations: One is long and the other is short. Format your response as JSON Object such that the keys will include "ClusterID + Cluster short explanation" and the values will include "Cluster short explanation"}}. 
Note that, we have tried to tune the prompts with a held-out validation set of cluster and explanations, and after manual inspection we finally chose the best prompt after prompt-engineering with 20 prompts. 
For each cluster, we obtain a short and an illustrated explanation.

\subsection{Prompt-2}
\label{appendix:promptB}
\textit{\enquote{Can you cluster these questions "+allquestions+ "into "+ K +" clusters, such that a single question should not be placed in more than one cluster and each of the clusters are semantically coherent, also provide a short and a long explanation for each cluster being generated. Format your response as a list of JSON Objects where the keys can be "ClusterID", "ClusterName", "ClusterExplanation" and "ClusterContent" where clusterid should start as Cluster1 and "ClusterExplanation" is the free-form explanation of the cluster description, "ClusterName" is the abbreviated form of the description and "ClusterContent" contains the questions to be grouped together in the corresponding cluster.}}

Note that, we have tried to tune the prompts with a held-out validation set of cluster and explanations, and after manual inspection we finally chose the best prompt after prompt-engineering with 10 prompts. 
Here also, for each cluster, we have a short and an illustrated explanations. 

\subsection{Reclustering based on user edits from Document-Level Cluster View}
\label{appendix:inference}
In Figure \ref{fig:inferexp}, we observe that the user has rearranged some questions related to heparin side effects into an existing cluster 6 which was mapped to "Cause" Slot, and the existing explanation is shown in blue color. However, after the edits, new explanation is shown to the user on pressing the "Infer Explanations after Edits" button. It helps the users know whether their edits have changed the goal of an existing cluster before the user takes a decision of pressing "Refine and Lock" in order to permanently save his edits into the system. Therefore, our system also lets the users know about the new explanation caused by his modification, and asks him if he wants to recluster based on the changes by asking a Yes/No question. If the user presses yes, then reclustering happens on the new explanations, otherwise not.

\end{document}